%% file: main.tex
\PassOptionsToPackage{table}{xcolor}
\documentclass[11pt]{article}

\usepackage{acl}

\usepackage{times}
\usepackage{latexsym}
\usepackage[T1]{fontenc}
\usepackage[utf8]{inputenc}
\usepackage{microtype}
\usepackage{inconsolata}

\usepackage{graphicx}
\usepackage{subcaption}
\usepackage{amsmath}
\usepackage{amssymb}
\usepackage{mathtools}
\usepackage{bm}
\usepackage{booktabs}
\usepackage{multirow}
\usepackage{makecell}
\usepackage{pifont}
\usepackage{array}
\usepackage{tabularx}
\usepackage[capitalize,noabbrev]{cleveref}
\usepackage[most]{tcolorbox}
\usepackage{fvextra}
\usepackage{fancyvrb}

\newcommand{\cmark}{\ding{51}}
\newcommand{\xmark}{\ding{55}}

\newcolumntype{Y}{>{\centering\arraybackslash}X}

\definecolor{mydarkgreen}{RGB}{0,100,0}
\tcbuselibrary{breakable}
\newtcolorbox{mycustombox}[1]{
  enhanced,
  colback=black!5!white,
  colframe=black!75!white,
  boxrule=0.4pt,
  coltitle=white,
  title=#1,
  titlerule=0.4pt,
  fontupper=\small,
  fonttitle=\small,
  before upper={\par\smallskipamount},
  breakable,
  left=3mm,
  right=3mm,
  top=2mm,
  bottom=2mm,
  boxsep=1mm,
}
\DefineVerbatimEnvironment{verbatim}{Verbatim}{
  breaklines=true,
  breakanywhere=true,
  breaksymbolleft=,
  breaksymbolright=
}

\title{$R^2$-dLLM: Accelerating Diffusion Large Language Models via Spatio-Temporal Redundancy Reduction}

\author{
  \textbf{Zhenbang Du$^{1}$, Kejing Xia$^{1}$, Xinrui Zhong$^{1}$, Yonggan Fu$^{2}$,} \\
  \textbf{Nicolai Oswald$^{2}$, Binfei Ji$^{1}$, Brucek Khailany$^{2}$, Pavlo Molchanov$^{2}$,} \\
  \textbf{Yingyan (Celine) Lin$^{1}$} \\
  $^{1}$Georgia Institute of Technology,
  $^{2}$NVIDIA
}

\begin{document}
\maketitle

\begin{abstract}
Diffusion Large Language Models (dLLMs) have emerged as a promising alternative to autoregressive generation by enabling parallel token prediction. However, practical dLLM decoding still suffers from high inference latency, which limits deployment. In this work, we observe that a substantial part of this inefficiency comes from recurring redundancy in the decoding process, including spatial redundancy caused by confidence clusters and positional ambiguity, and temporal redundancy caused by repeatedly remasking predictions that have already stabilized. Motivated by these patterns, we propose \textbf{$\bm{R^2}$-dLLM}, a unified framework for reducing decoding redundancy from both inference and training perspectives. At inference time, we introduce training-free decoding rules that aggregate local confidence and token predictions, and finalize temporally stable tokens to avoid redundant decoding steps. We further propose a redundancy-aware supervised fine-tuning pipeline that aligns the model with efficient decoding trajectories and reduces reliance on manually tuned thresholds. Experiments demonstrate that $R^{2}$-dLLM consistently reduces the number of decoding steps by up to 88\% compared to existing decoding strategies, while maintaining competitive generation quality across different models and tasks. These results validate that decoding redundancy is a central bottleneck in dLLMs, and that explicitly reducing it yields substantial practical efficiency gains. Our code and models are available at \href{https://github.com/GATECH-EIC/R2-dLLM}{https://github.com/GATECH-EIC/R2-dLLM}.
\end{abstract}

\input{Sections/1-Introduction}
\input{Sections/2-Related_works}

\input{Sections/2.5-Preliminaries}

\input{Sections/3-Methods}
\input{Sections/4-Experiments}

\input{Sections/5-Conclusion}

\section*{Limitations}

This work studies spatio-temporal redundancy mainly on LLaDA-Instruct-8B, LLaDA-1.5, and Dream-v0-Instruct-7B, using math and code benchmarks. Redundancy behavior may differ for other models or open-ended tasks. The training-free variant still depends on thresholds, while redundancy-aware SFT introduces offline data collection and training cost.

\bibliography{main}

\clearpage
\input{Sections/Supp}

\end{document}

%% file: Sections/1-Introduction.tex
\section{Introduction}

\begin{figure}[t]
    \centering
    \includegraphics[width=0.9\linewidth]{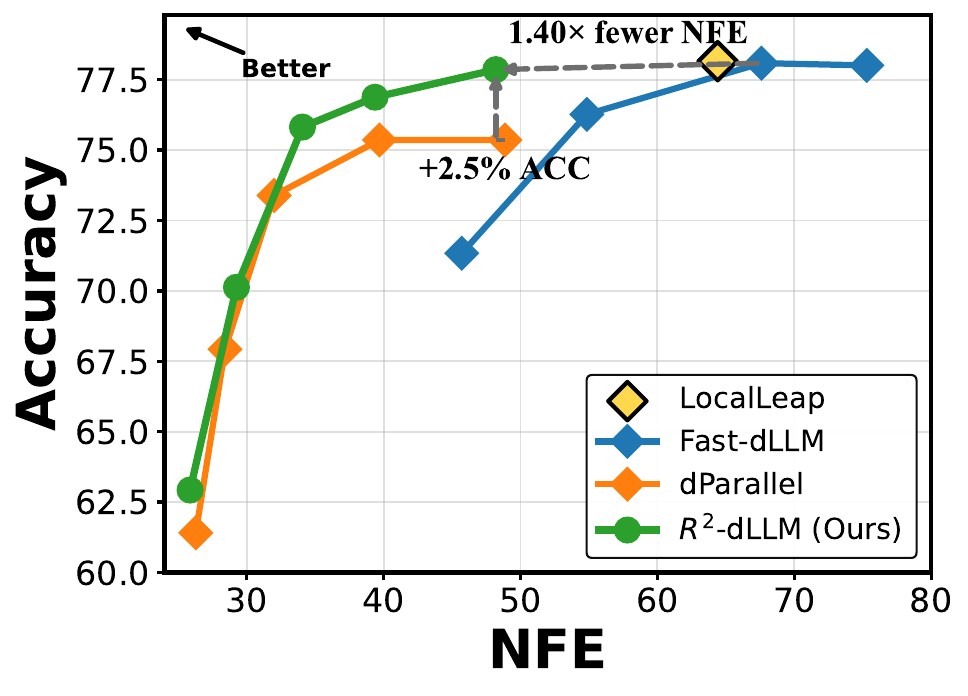}
        \caption{Benchmarking the accuracy versus the Number of Function Evaluations (NFE) trade-offs between our $R^{2}$-dLLM and SOTA dLLM acceleration methods on the GSM8K dataset based on the LLaDA-Instruct-8B model.}
    \label{fig:tradeoff}
    \vspace{-12pt}
\end{figure}

Diffusion Large Language Models (dLLMs) have recently emerged as a compelling alternative to the dominant AutoRegressive (AR) paradigm \cite{brown2020language,ouyang2022training}. Unlike AR models, which are constrained by a sequential and token-by-token generation bottleneck, dLLMs leverage bidirectional attention and iterative decoding to predict multiple tokens in parallel ~\cite{nie2025large, ye2025dream}. This non-autoregressive modeling capability theoretically allows for significantly higher decoding throughput and higher hardware utilization for low-batch scenarios, offering a potential direction toward more efficient large-scale text generation.

However, translating this theoretical benefit into practical speedups remains a significant challenge. Despite their generation flexibility, early-stage dLLMs often suffer from higher inference latency than AR models. Recently, many pioneering works are proposed to address this issue, such as Fast-dLLM ~\cite{wu2025fast}, which introduces confidence-based parallel decoding to accelerate token unmasking; dLLM-Cache ~\cite{liu2025dllm}, which implements specialized KV-Caching mechanisms to eliminate unnecessary attention computations across decoding steps; and D2F ~\cite{wang2025diffusion}, which adopts a semi-autoregressive attention mechanism instead of fully bidirectional attention to minimize unnecessary attention computations across different decoding blocks. While these approaches reduce computation from architectural or system perspectives, practical dLLM decoding still exhibits inefficiencies within the decoding trajectory itself, namely \textit{redundancy in the decoding dynamics}.

In this paper, we study two forms of redundancy in dLLM decoding under a unified spatio-temporal view: 1) \textbf{Spatial Redundancy}: Unlike AR models that generate tokens at fixed positions, dLLMs must jointly determine token content and positions. This leads to two common patterns. First, dLLMs often produce \emph{confidence clusters}, where multiple neighboring tokens simultaneously reach high confidence \cite{kong2025accelerating}. These tokens are strongly correlated and could be decoded together, but mainstream existing decoding strategies still process them independently, leading to redundant decoding steps. Second, positional uncertainty causes \emph{token clusters}, where the same token is repeatedly predicted at multiple adjacent positions. The model then needs extra decoding steps to finalize the exact position. 2) \textbf{Temporal Redundancy}: During iterative decoding, many tokens converge early and their predictions remain unchanged for multiple steps \cite{shen2025improving}. However, existing strategies continue to remask and re-decode these tokens. This leads to unnecessary computation on already confident predictions.

To alleviate these issues, we propose \underline{R}edundancy-\underline{R}educed-dLLM (\textbf{$\bm{R^2}$-dLLM}), a unified framework that targets both spatial and temporal redundancies in dLLM decoding. We first introduce a \textit{training-free} decoding strategy that directly reduces redundancy during inference. To address spatial redundancy, where models struggle with token positioning, we apply a local confidence and token aggregation mechanism to stabilize consecutive confidence and token clusters. To handle temporal redundancy, we further introduce a multi-step consistency check that finalizes tokens with stable predictions across consecutive steps, thereby avoiding unnecessary remasking. Beyond accelerating the decoding process, this training-free strategy also provides a practical way to measure redundancy scores along generation trajectories. Building on this observation, we further propose a \textit{redundancy-aware supervised fine-tuning} pipeline that aligns the model with redundancy-less decoding behaviors. Specifically, we construct the training dataset by filtering generation trajectories with minimal redundancy scores, enabling the model to directly learn redundancy reduction patterns and reducing its dependence on manually tuned thresholds required by training-free decoding strategies. Our $R^{2}$-dLLM, as shown in Figure~\ref{fig:tradeoff}, achieves a favorable trade-off between decoding efficiency and generation quality by reducing redundancy in the decoding process.

In summary, our contributions are as follows:

\begin{itemize}
    \item We present a unified view of spatio-temporal redundancy in dLLM decoding, covering locally adjacent high-confidence predictions, token-cluster positional ambiguity, and repeated decoding of already stable tokens.

    \item We propose $R^{2}$-dLLM, a unified framework that reduces redundancy from both inference and training perspectives, including training-free decoding strategies to alleviate inherent spatial and temporal redundancy, and a redundancy-aware supervised fine-tuning pipeline that aligns the model with efficient decoding trajectories.

    \item Extensive experiments across representative benchmarks demonstrate that $R^{2}$-dLLM significantly reduces the Number of Function Evaluations (NFE) and improves dLLM inference efficiency, while maintaining competitive generation quality.
\end{itemize}

%% file: Sections/2-Related_works.tex
\section{Related Works}

\subsection{Diffusion Language Models}

Traditional AR language models generate text tokens from left to right \cite{brown2020language,ouyang2022training}. While they can produce high-quality text, their decoding throughput is limited by sequential generation. Recently, diffusion probabilistic models have been extended to discrete text generation \cite{austin2021structured, chen2022analog, gulrajani2023likelihood}. dLLMs generate text by iteratively predicting and unmasking tokens, which enables multiple tokens to be updated in a single forward pass.
LLaDA \cite{nie2025large} scales diffusion language models to 8B parameters and trains them from scratch. Dream 7B \cite{ye2025dream} instead initializes from a pretrained AR model (Qwen-2.5 7B) \cite{yang2025qwen3}. While these efforts advance dLLMs, redundancy during decoding remains prevalent, and the potential parallelism of dLLM decoding is still not fully exploited.

\subsection{dLLM Acceleration}
Recent work has explored accelerating dLLM decoding from multiple angles. 
Caching-based methods reduce computation by reusing KV caches across decoding steps \cite{ma2025dkv,liu2025dllm,hu2025accelerating}. Fast-dLLM accelerates decoding by using a predefined confidence threshold to unmask multiple tokens in parallel \cite{wu2025fast}. 
D2F further reduces inference cost by replacing fully bidirectional attention with a semi-autoregressive attention mechanism, avoiding unnecessary attention computation across decoding blocks \cite{wang2025diffusion}. Quantization-based approaches improve efficiency by compressing dLLMs to low-precision formats \cite{xu2025dllmquant}.
Recent methods also improve parallel decoding through dependency-aware or adaptive token scheduling \cite{luo2026dawn,luo2026divide,chen2026dmax}.
Beyond reducing raw computation, several methods target more stable decoding dynamics. CreditDecoding \cite{wang2025creditdecoding} accumulates token-level consistency across steps. dParallel \cite{chen2025dparallel} applies certainty-forcing distillation, and LocalLeap \cite{kong2025accelerating} groups locally consistent tokens to reduce repeated updates. 
Despite these advances, existing approaches often optimize only specific components of the decoding pipeline and depend on manually tuned hyperparameters, leading to suboptimal accuracy--efficiency trade-offs.

%% file: Sections/2.5-Preliminaries.tex
\section{Preliminaries}
\textbf{Diffusion Language Models.}
dLLMs formulate text generation as a discrete denoising process over token sequences.
Given a clean sequence $x_0 = (x_{0,1}, \ldots, x_{0,L})$, we define a corruption process parameterized by a noise level $t \in (0,1]$:
each token is independently replaced by a special mask token $M$ with probability $t$, producing a partially observed sequence $x_t$.
When $t=1$, all tokens are masked; when $t$ is close to $0$, $x_t$ is close to $x_0$.
A model $p_\theta$ is trained to predict the original tokens at masked positions conditioned on $x_t$.

\textbf{Training Objective.}
For each training example, we sample $t$ and construct $x_t$ by independent masking.
We optimize the denoising objective
\begin{equation}
\begin{aligned}
\mathcal{L}(\theta)
= -\mathbb{E}_{t,x_0,x_t}
\bigg[
\frac{1}{t}\sum_{i=1}^{L}\mathbb{I}[x_{t,i}=M] \\
\cdot \log p_\theta(x_{0,i}\mid x_t)
\bigg].
\end{aligned}
\end{equation}
For conditional generation, we apply masking only to response tokens while keeping the prompt fixed.

\paragraph{Inference Phase.}
Given a prompt $x_p$, dLLMs generate a response $r_0 \sim p_\theta(r_0 \mid x_p)$ via an iterative unmasking--remasking procedure. Decoding starts from a fully masked response sequence. At each step, the model predicts token distributions for all currently masked positions in parallel; a subset of positions is then finalized (unmasked), while the remaining positions stay masked (or are remasked) for subsequent steps.

Vanilla diffusion decoding typically finalizes the top-$k$ (e.g., $k=1$ in LLaDA \cite{nie2025large}) predictions based on confidence scores at each step. Fast-dLLM instead finalizes all predictions whose confidence exceeds a predefined threshold $\tau$, enabling more tokens to be unmasked per step and thereby reducing the number of decoding steps. While these strategies improve inference efficiency, they still induce redundancy along the decoding trajectory, which we analyze in subsequent sections.

%% file: Sections/3-Methods.tex
\section{Redundancies in dLLMs}
\label{sec:redundancy}
In this section, we analyze how redundancy arises during dLLM decoding and introduce simple methods to alleviate spatial and temporal redundancy.

\begin{figure}[h]
    \centering
    \includegraphics[width=0.99\linewidth]{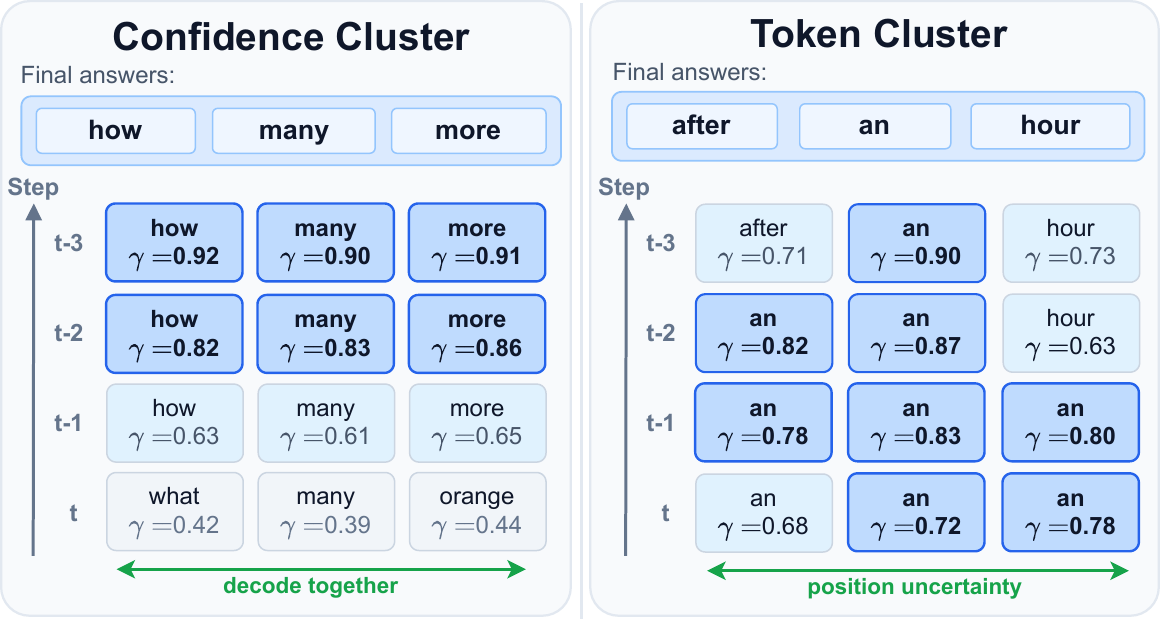}
    \caption{An illustration of spatial redundancy.}
    \label{fig:spatial}
\end{figure}

\begin{figure*}[h]
    \centering
    \includegraphics[width=\linewidth]{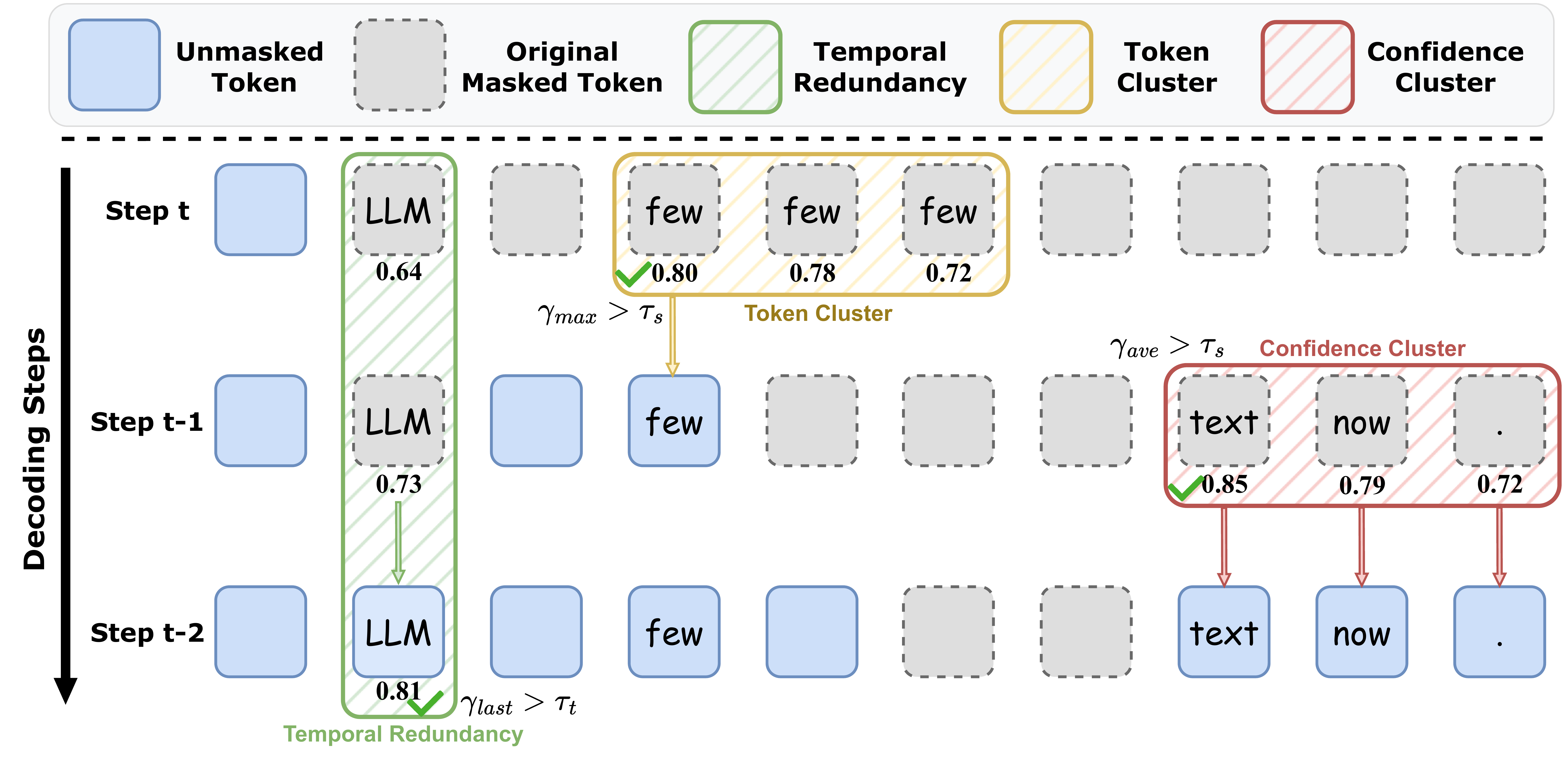}
    \caption{Overview of the proposed training-free redundancy reduction strategy during diffusion decoding.}
    \label{fig:trainingfree}
    \vspace{-10pt}
\end{figure*}

\subsection{Spatial Redundancy}

Unlike AR models that decode tokens in a fixed order, dLLMs generate tokens in a random order. While this enables parallel token prediction, it also introduces uncertainty over token positions. As a result, dLLMs must jointly determine both the token content and positions, which causes spatial redundancy during decoding.

One common form of spatial redundancy is the \emph{confidence clusters}. Here, multiple neighboring tokens reach relatively high confidence at the same decoding step and could be decoded together, as illustrated in Figure \ref{fig:spatial}. Existing decoding strategies mostly process such tokens independently, which delays their decoding and introduces unnecessary steps.

Another form of spatial redundancy is the \emph{token cluster} phenomenon, as shown in Figure \ref{fig:spatial}. In particular, the same token is repeatedly predicted at multiple adjacent positions, where the model has correctly identified the token content but is still uncertain about its exact position. The model then requires multiple steps to resolve this positional ambiguity.

To resolve spatial redundancy, we apply a local aggregation rule guided by confidence statistics.
\paragraph{Confidence Cluster Aggregation.}
When multiple neighboring tokens reach relatively high confidence at the same step, we treat them as a confidence cluster.
For a local window $\mathcal{W}$, we compute the average confidence
\begin{equation}
\gamma_{\text{ave}} = \frac{1}{|\mathcal{W}|} \sum_{i \in \mathcal{W}} \gamma_i,
\end{equation}
where $\gamma_i$ denotes the confidence score at position $i$.
If $\gamma_{\text{ave}} > \tau_s$ (a prespecified threshold), all tokens in $\mathcal{W}$ are decoded simultaneously.

\paragraph{Token Cluster Aggregation.}
When the same token is predicted at multiple adjacent positions (e.g., $\geq2$ positions), we treat them as a token cluster.
Given a cluster $\mathcal{C} = \{i_1,\dots,i_k\}$, we select the position with the maximum confidence,
\begin{equation}
i^* = \arg\max_{i \in \mathcal{C}} \gamma_i .
\end{equation}
If $\gamma_{i^*} > \tau_s$, the token is directly decoded at position $i^*$, resolving the positional ambiguity.

\subsection{Temporal Redundancy}

\begin{figure}[h]
    \centering
    \includegraphics[width=0.90\linewidth]{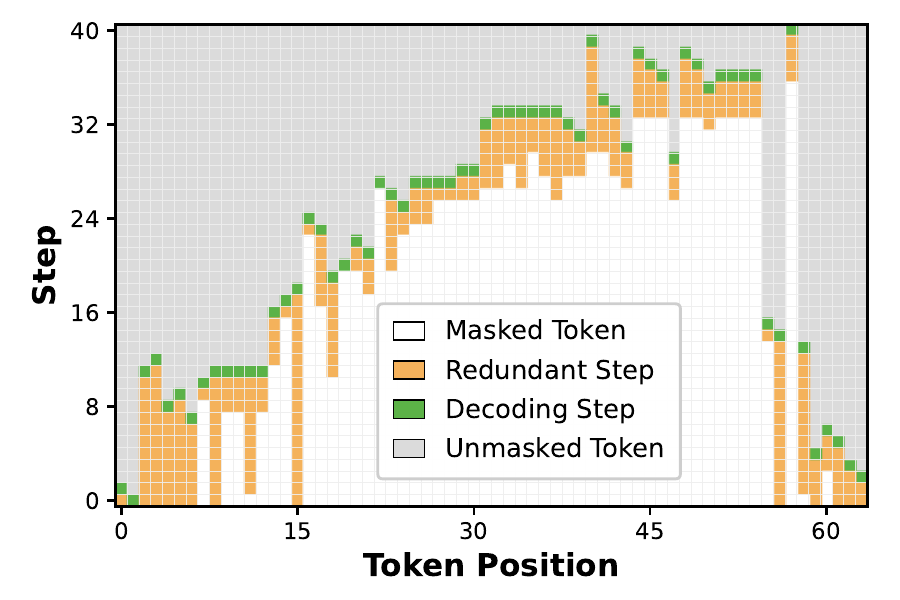}
    \caption{An illustration of temporal redundancy, where a ``redundant step'' denotes a token that matches the final output but is not finalized.}
    \label{fig:temporal}
        \vspace{-10pt}
\end{figure}

dLLMs generate text through an iterative unmasking--remasking process. At each decoding step, the model predicts tokens at masked positions and only finalizes a subset of them based on the prediction confidence, while the remaining tokens are remasked and unmasked in later steps. This can introduce temporal redundancy.

In practice, many tokens converge early during decoding. Their predicted token remains unchanged across multiple consecutive steps, as shown in Figure \ref{fig:temporal}. However, existing decoding strategies mostly remask these tokens until the final decoding step or until the corresponding confidence exceeds a sufficiently high threshold (e.g., 0.9). As a result, the model repeatedly decodes already stable tokens, leading to unnecessary computation and decoding steps.

To reduce temporal redundancy, we introduce a multi-step consistency check to finalize early-converged tokens. Specifically, for a given position $i$, if the predicted token remains unchanged for $m$ consecutive decoding steps, and the confidence at the last step satisfies
\begin{equation}
\gamma_{i}^{\text{last}} > \tau_t ,
\end{equation}
then this token is finalized. In practice, we find that a small value of $m$ (e.g., $m=3$) works well across different models and tasks. This rule allows stable tokens to be decoded earlier and avoids redundant decoding.

Together, these training-free rules explicitly reduce both spatial and temporal redundancy during the decoding process, as illustrated in Figure \ref{fig:trainingfree}.

\section{Redundancy-aware Supervised Fine-Tuning}

The aforementioned training-free decoding strategies effectively reduce redundancy during inference. To further enhance the achievable performance, we introduce a redundancy-aware supervised fine-tuning approach that encourages the model to internalize efficient decoding behaviors during training, which at the same time can reduce its dependence on heuristic rules at inference time.

\begin{figure}[h]
    \centering
    \includegraphics[width=\linewidth]{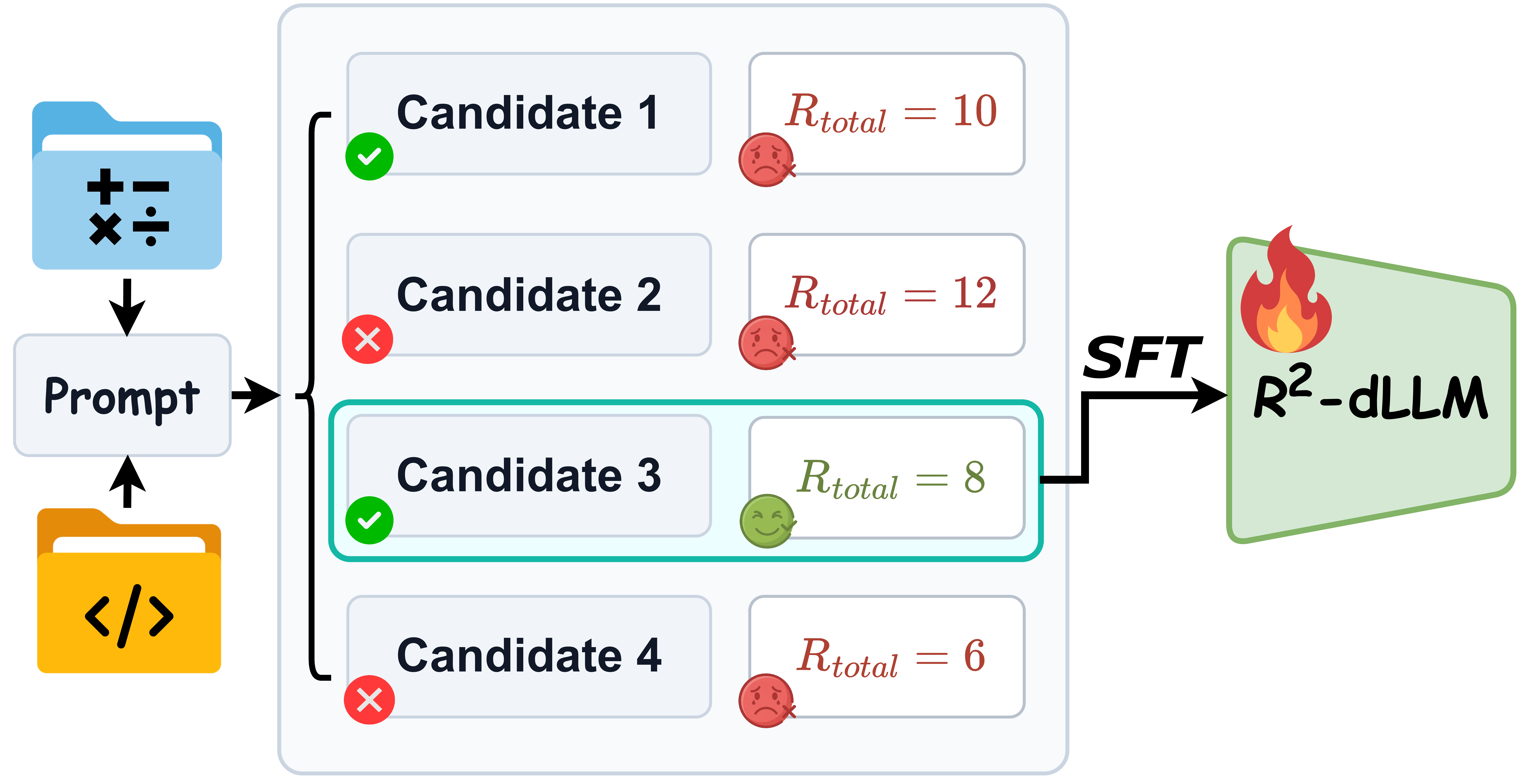}
    \caption{Redundancy-aware training dataset collection: For each prompt, candidate responses with correct answers and the lowest redundancy score $R_{\text{total}}$ are selected.}
    \label{fig:dataset_collect}
        \vspace{-10pt}
\end{figure}

\subsection{Training Dataset Collection}

We construct a redundancy-aware training dataset by explicitly selecting decoding trajectories with minimal redundancy, as illustrated in Figure~\ref{fig:dataset_collect}. Specifically, for each prompt, we first generate multiple candidate responses using a pretrained dLLM. During generation, we track redundancy statistics along each decoding trajectory using the training-free rules introduced in Section~\ref{sec:redundancy}. 
Each triggered spatial aggregation contributes one point of spatial redundancy $R_s$, and each triggered temporal consistency check contributes one point of temporal redundancy $R_t$. 
The overall redundancy score of a candidate response is defined as $R_{\text{total}} = R_s + R_t$.

Given a set of candidate responses for the same prompt, we select the response with the lowest $R_{\text{total}}$, which corresponds to a decoding trajectory that converges faster and exhibits fewer redundancies. The selected response is then paired with the original prompt to form a supervised training example. Repeating this process across prompts yields a training dataset that captures efficient diffusion decoding behavior.

\begin{table*}[t]
  \centering
  \small
  \setlength{\tabcolsep}{6pt}
  \renewcommand{\arraystretch}{1.12}

  \newcommand{\speedup}[1]{$_{\textcolor{green!60!black}{\scriptsize\downarrow #1\times}}$}
  \newcommand{\nfered}[1]{$_{\textcolor{green!60!black}{\scriptsize\downarrow #1\%}}$}
  \newcolumntype{C}[1]{>{\centering\arraybackslash}m{#1}}

  \begin{tabular}{c c C{1.5cm} C{1.20cm} C{1.15cm} C{1.5cm} C{1.20cm} C{1.15cm}}
    \toprule
    \multirow{2}{*}{\textbf{Task}} &
    \multirow{2}{*}{\textbf{Method}} &
    \multicolumn{3}{c}{\textbf{LLaDA-Instruct-8B}} &
    \multicolumn{3}{c}{\textbf{Dream-v0-Instruct-7B}} \\
    \cmidrule(lr){3-5}\cmidrule(lr){6-8}
    & &
    \textbf{Latency$\downarrow$} & \textbf{NFE$\downarrow$} & \textbf{ACC$\uparrow$} &
    \textbf{Latency$\downarrow$} & \textbf{NFE$\downarrow$} & \textbf{ACC$\uparrow$} \\
    \midrule

    \multirow{6}{*}{\makecell[c]{GSM8K\\(5-shot)}}
      & Vanilla
        & 13.1 & 256.0 & 77.03
        & 10.4 & 256.0 & 78.77 \\
      & Fast-dLLM
        & 2.7\speedup{4.9} & 86.4\nfered{66} & 78.01
        & 1.7\speedup{6.1} & 63.6\nfered{75} & 76.57 \\
      & LocalLeap
        & 2.3\speedup{5.7} & 64.4\nfered{74} & 78.16
        & 1.4\speedup{7.4} & 61.0\nfered{76} & 76.19 \\
      & dParallel
        & 1.4\speedup{9.4} & 39.7\nfered{84} & 75.36
        & 1.3\speedup{8.0} & 38.7\nfered{84} & 78.47 \\
      & \cellcolor{gray!15}\textbf{$\bm{R^2}$-dLLM (F)}
        & \cellcolor{gray!15}2.3\speedup{5.7} & \cellcolor{gray!15}63.1\nfered{75} & \cellcolor{gray!15}77.56
        & \cellcolor{gray!15}1.4\speedup{7.4} & \cellcolor{gray!15}53.3\nfered{79} & \cellcolor{gray!15}75.51 \\
      & \cellcolor{gray!15}\textbf{$\bm{R^2}$-dLLM (T)}
        & \cellcolor{gray!15}1.7\speedup{7.7} & \cellcolor{gray!15}51.9\nfered{79} & \cellcolor{gray!15}77.86
        & \cellcolor{gray!15}1.3\speedup{8.0} & \cellcolor{gray!15}44.9\nfered{82} & \cellcolor{gray!15}80.82 \\
    \midrule

    \multirow{6}{*}{\makecell[c]{MATH\\(4-shot)}}
      & Vanilla
        & 9.7 & 256.0 & 33.20
        & 8.3 & 256.0 & 38.08 \\
      & Fast-dLLM
        & 3.3\speedup{2.9} & 107.8\nfered{57} & 32.60
        & 2.4\speedup{3.5} & 90.3\nfered{64} & 37.58 \\
      & LocalLeap
        & 2.8\speedup{3.5} & 85.3\nfered{66} & 32.18
        & 1.9\speedup{4.4} & 81.5\nfered{68} & 37.18 \\
      & dParallel
        & 1.7\speedup{5.7} & 53.8\nfered{78} & 30.22
        & 1.6\speedup{5.2} & 57.3\nfered{77} & 35.94 \\
      & \cellcolor{gray!15}\textbf{$\bm{R^2}$-dLLM (F)}
        & \cellcolor{gray!15}2.9\speedup{3.3} & \cellcolor{gray!15}79.7\nfered{68} & \cellcolor{gray!15}32.26
        & \cellcolor{gray!15}1.7\speedup{4.9} & \cellcolor{gray!15}76.3\nfered{70} & \cellcolor{gray!15}37.56 \\
      & \cellcolor{gray!15}\textbf{$\bm{R^2}$-dLLM (T)}
        & \cellcolor{gray!15}2.1\speedup{4.6} & \cellcolor{gray!15}68.5\nfered{73} & \cellcolor{gray!15}32.60
        & \cellcolor{gray!15}1.8\speedup{4.6} & \cellcolor{gray!15}66.1\nfered{74} & \cellcolor{gray!15}36.88 \\
    \midrule

    \multirow{6}{*}{\makecell[c]{HumanEval\\(0-shot)}}
      & Vanilla
        & 5.8 & 256.0 & 40.24
        & 4.7 & 256.0 & 57.93 \\
      & Fast-dLLM
        & 2.5\speedup{2.3} & 90.1\nfered{64} & 36.59
        & 1.7\speedup{2.8} & 80.5\nfered{68} & 57.93 \\
      & LocalLeap
        & 2.1\speedup{2.8} & 69.8\nfered{72} & 35.98
        & 1.4\speedup{3.4} & 75.0\nfered{70} & 53.66 \\
      & dParallel
        & 1.1\speedup{5.3} & 37.9\nfered{85} & 33.53
        & 1.1\speedup{4.3} & 50.4\nfered{80} & 53.66 \\
      & \cellcolor{gray!15}\textbf{$\bm{R^2}$-dLLM (F)}
        & \cellcolor{gray!15}2.3\speedup{2.5} & \cellcolor{gray!15}81.7\nfered{68} & \cellcolor{gray!15}35.98
        & \cellcolor{gray!15}1.3\speedup{3.6} & \cellcolor{gray!15}67.5\nfered{73} & \cellcolor{gray!15}53.05 \\
      & \cellcolor{gray!15}\textbf{$\bm{R^2}$-dLLM (T)}
        & \cellcolor{gray!15}1.7\speedup{3.4} & \cellcolor{gray!15}64.3\nfered{74} & \cellcolor{gray!15}36.59
        & \cellcolor{gray!15}0.9\speedup{5.2} & \cellcolor{gray!15}44.6\nfered{82} & \cellcolor{gray!15}54.27 \\
    \midrule

    \multirow{6}{*}{\makecell[c]{MBPP\\(3-shot)}}
      & Vanilla
        & 10.1 & 256.0 & 29.40
        & 7.9 & 256.0 & 61.00 \\
      & Fast-dLLM
        & 2.2\speedup{4.6} & 73.0\nfered{71} & 25.60
        & 1.0\speedup{7.9} & 35.3\nfered{86} & 53.20 \\
      & LocalLeap
        & 1.9\speedup{5.3} & 56.1\nfered{78} & 24.00
        & 0.9\speedup{8.8} & 36.3\nfered{85} & 51.40 \\
      & dParallel
        & 1.0\speedup{10.1} & 29.7\nfered{88} & 35.80
        & 0.8\speedup{9.9} & 26.6\nfered{89} & 48.20 \\
      & \cellcolor{gray!15}\textbf{$\bm{R^2}$-dLLM (F)}
        & \cellcolor{gray!15}1.9\speedup{5.3} & \cellcolor{gray!15}61.2\nfered{76} & \cellcolor{gray!15}25.60
        & \cellcolor{gray!15}0.8\speedup{9.9} & \cellcolor{gray!15}29.9\nfered{88} & \cellcolor{gray!15}52.60 \\
      & \cellcolor{gray!15}\textbf{$\bm{R^2}$-dLLM (T)}
        & \cellcolor{gray!15}1.3\speedup{7.7} & \cellcolor{gray!15}40.3\nfered{84} & \cellcolor{gray!15}37.40
        & \cellcolor{gray!15}0.8\speedup{9.9} & \cellcolor{gray!15}28.7\nfered{88} & \cellcolor{gray!15}54.00 \\
    \bottomrule
  \end{tabular}
    \caption{Efficiency and accuracy comparison under different decoding strategies (generation length is equal to 256). Latency speedup and NFE reduction over Vanilla is shown in green.}
  \label{tab:main_results}
   \vspace{-10pt}
\end{table*}

\subsection{Supervised Fine-Tuning}

To maintain consistency with the inference procedure, we adopt a semi-autoregressive training scheme for supervised fine-tuning \cite{chen2025dparallel}. Given a prompt--response pair from the collected dataset, we concatenate the prompt and response to form a full sequence $x_0$. During training, prompt tokens remain unmasked, while response tokens are divided into non-overlapping blocks of size $s$ (we use $s=32$ in all experiments, following existing dLLM work).

At each training step, we randomly sample a block index $i$. Meanwhile, blocks with indices smaller than $i$ are kept unmasked, while blocks with indices larger than $i$ remain fully masked. Within block $i$, we sample a binary mask $m \in \{0,1\}^{s}$ with masking probability $p_m$, where $m_j = 1$ indicates that token $j$ is masked. The model is trained to predict the masked tokens in the selected block, with the loss defined as
\begin{equation}
\begin{aligned}
&\mathcal{L}_{\text{SFT}}
= {}  \\
&-\mathbb{E}\!\bigg[
\frac{1}{s\times p_m}
\sum_{j=1}^{s}
\mathbf{1}[m_j = 1] \\
&\quad
\log p_\theta(x_{i,j} \mid x_{<i}, x_{i,\neg m})
\bigg]
\end{aligned}
\end{equation}
where $x_{i,\neg m}$ denotes the unmasked tokens in block $i$.

We also adopt a complementary sampling strategy~\cite{wu2025fast}. In particular, for each training example, we construct a complementary mask $\bar{m} = 1 - m$ and create two training views corresponding to $m$ and $\bar{m}$. This ensures that each token within a block is supervised once as masked and once as unmasked, which helps reduce overfitting and stabilizes training.

%% file: Sections/4-Experiments.tex
\section{Experiments}

\begin{table*}[t]
  \centering
  \small
  \setlength{\tabcolsep}{5pt}
  \renewcommand{\arraystretch}{1.12}

  \newcommand{\nfeann}[2]{    \multicolumn{1}{c}{#1\rlap{\hspace{0em}\textcolor{green!60!black}{\scriptsize$\downarrow$#2\%}}}  }

  \begin{tabular*}{0.75\textwidth}{@{\extracolsep{\fill}}ccccccccc}
\toprule
\multirow{2}{*}{\textbf{Method}} &
\multicolumn{2}{c}{\textbf{GSM8K}} &
\multicolumn{2}{c}{\textbf{MATH}} &
\multicolumn{2}{c}{\textbf{HumanEval}} &
\multicolumn{2}{c}{\textbf{MBPP}} \\
\cmidrule(lr){2-3}
\cmidrule(lr){4-5}
\cmidrule(lr){6-7}
\cmidrule(lr){8-9}
&
\textbf{NFE $\downarrow$} & \textbf{ACC $\uparrow$} &
\textbf{NFE $\downarrow$} & \textbf{ACC $\uparrow$} &
\textbf{NFE $\downarrow$} & \textbf{ACC $\uparrow$} &
\textbf{NFE $\downarrow$} & \textbf{ACC $\uparrow$} \\
\midrule
Fast-dLLM
  & 84.5 & 79.81
  & 106.6 & 32.54
  & 104.7 & 36.59
  & 48.5 & 34.40 \\
$\bm{R^2}$\textbf{-dLLM}
  & \nfeann{50.5}{40} & 79.23
  & \nfeann{72.5}{32} & 32.40
  & \nfeann{81.8}{21} & 37.20
  & \nfeann{38.0}{21} & 37.00 \\
\bottomrule
\end{tabular*}
 \caption{Efficiency and accuracy comparison between Fast-dLLM and $R^2$-dLLM on LLaDA-1.5. NFE reduction is computed relative to Fast-dLLM, shown in green.}
  \label{tab:llada1.5}
\end{table*}

\begin{table*}[!h]
  \centering
  \small
  \setlength{\tabcolsep}{4pt}
  \renewcommand{\arraystretch}{1.12}
 
  \resizebox{\textwidth}{!}{%
  \begin{tabular}{ccc cccccccc}
    \toprule
    \multicolumn{3}{c}{\textbf{Redundancy Type}} &
    \multicolumn{2}{c}{\textbf{GSM8K}} &
    \multicolumn{2}{c}{\textbf{MATH}} &
    \multicolumn{2}{c}{\textbf{HumanEval}} &
    \multicolumn{2}{c}{\textbf{MBPP}} \\
    \cmidrule(lr){1-3}
    \cmidrule(lr){4-5}
    \cmidrule(lr){6-7}
    \cmidrule(lr){8-9}
    \cmidrule(lr){10-11}
    \textbf{Spa. (Conf.)} & \textbf{Spa. (Token.)} & \textbf{Temp.} &
    \textbf{NFE $\downarrow$} & \textbf{ACC $\uparrow$} &
    \textbf{NFE $\downarrow$} & \textbf{ACC $\uparrow$} &
    \textbf{NFE $\downarrow$} & \textbf{ACC $\uparrow$} &
    \textbf{NFE $\downarrow$} & \textbf{ACC $\uparrow$} \\
    \midrule
    \xmark & \xmark & \xmark &
    86.4 & 78.01 &
    107.8 & 32.60 &
    90.1 & 36.59 &
    73.0 & 25.60 \\
    \midrule
    \cmark &  &  &
    69.6 & 77.33 &
    89.4 & 32.62 &
    79.1 & 36.59 &
    61.5 & 25.40 \\
     & \cmark &  &
    80.3 & 78.32 &
    104.2 & 32.20 &
    86.5 & 35.37 &
    72.5 & 25.40 \\
     &  & \cmark &
    70.0 & 77.51 &
    90.7 & 32.40 &
    88.8 & 35.98 &
    63.2 & 25.80 \\
    \midrule
    \cmark & \cmark & \cmark &
    63.1 & 77.56 &
    79.7 & 32.26 &
    81.7 & 35.98 &
    61.2 & 25.60 \\
    \bottomrule
  \end{tabular}}
   \caption{Ablation of training-free redundancy reduction components on LLaDA.}
  \label{tab:training_free_ablation}
  \vspace{-10pt}
\end{table*}

\subsection{Experiment Settings}

\textbf{Models and datasets.}
We evaluate $R^2$-dLLM on two representative diffusion language models, LLaDA-Instruct-8B~\cite{nie2025large} and Dream-v0-Instruct-7B~\cite{ye2025dream}. We conduct experiments on four benchmarks, including two math reasoning datasets, GSM8K~\cite{cobbe2021training} and MATH~\cite{hendrycks2021measuring}, and two code generation datasets, HumanEval~\cite{chen2021evaluating} and MBPP~\cite{austin2021program}. We follow common settings and use 5-shot prompting for GSM8K, 4-shot for MATH, 0-shot for HumanEval, and 3-shot for MBPP. For evaluation metrics, we report averaged latency (second), the Number of Function Evaluations (NFE), and accuracy (ACC).

\textbf{Baselines.}
We compare our method with several baselines to evaluate both efficiency and performance, including Vanilla dLLM, which uses Top-1 confidence decoding, Fast-dLLM~\cite{wu2025fast}, which decodes tokens whose confidence is above a fixed threshold (0.9 by default),  LocalLeap~\cite{kong2025accelerating}, which decodes neighboring tokens with similar confidence jointly, and dParallel~\cite{chen2025dparallel}, which applies a distillation-based method to increase decoding confidence. All methods except Vanilla use the dual KV cache mechanism introduced in Fast-dLLM~\cite{wu2025fast}. For all experiments, the generation length is set to 256, and the decoding block size is set to 32.

We construct training data by sampling prompts from open-source datasets. For both models, prompts are sampled from the GSM8K training set~\cite{cobbe2021training}, PRM12K~\cite{lightman2023let}, a subset of the Numina-Math dataset~\cite{li2024numinamath}, a subset of the AceCode dataset~\cite{le2022coderl}, and a subset of the Big-Math CN\_K12 dataset~\cite{albalak2025big}. All training and inference are run on four NVIDIA H200 141 GB GPUs, and the latency (second) is measured by one NVIDIA H200 141 GB GPU. Detailed settings are provided in the appendix.

\subsection{Main Results}

Table~\ref{tab:main_results} reports the efficiency and accuracy comparison across different decoding strategies, measured by latency, NFE, and task accuracy. $R^2$-dLLM (F) represents only applying our training-\underline{f}ree redundancy reduction, and $R^2$-dLLM (T) utilizes SF\underline{T} to reduce redundancy. Overall, $R^2$-dLLM achieves consistent latency speedups and NFE reductions while maintaining competitive or improved accuracy on both LLaDA and Dream.

For LLaDA, on GSM8K, $R^2$-dLLM (T) reduces latency from 13.1\,s to 1.7\,s, corresponding to a $7.7\times$ speedup over Vanilla decoding.
This is accompanied by a reduction in NFE from 256 to 51.9, while slightly improving accuracy from 77.03 to 77.86.
On MATH, $R^2$-dLLM (T) achieves a $4.6\times$ latency speedup (9.7\,s $\rightarrow$ 2.1\,s) and reduces NFE to 68.5.
On HumanEval, latency is reduced from 5.8\,s to 1.7\,s ($3.4\times$), with NFE decreasing to 64.3 and accuracy comparable to Fast-dLLM.
On MBPP, $R^2$-dLLM (T) attains a $7.7\times$ latency speedup and reduces NFE from 256 to 40.3, while improving accuracy from 29.40 to 37.40.

For Dream, $R^2$-dLLM yields consistent improvements.
On MATH, latency is reduced from 8.3\,s to 1.8\,s ($4.6\times$), with NFE decreasing to 66.1 and accuracy remaining competitive.
On HumanEval, $R^2$-dLLM (T) achieves a $5.2\times$ latency speedup (4.7\,s $\rightarrow$ 0.9\,s) and reduces NFE to 44.6, while maintaining accuracy.

Although dParallel often achieves lower latency and NFE, it incurs noticeable accuracy degradation on some tasks.
In contrast, $R^2$-dLLM consistently delivers strong latency speedups with a more favorable accuracy--efficiency trade-off across all tasks.

Overall, these results demonstrate that $R^2$-dLLM achieves substantial end-to-end latency acceleration by reducing redundancy while preserving generation quality.

\subsection{Performance on dLLM Variant}

To evaluate the generality of $R^2$-dLLM, we apply it to LLaDA-1.5~\cite{zhu2025llada}, a reinforcement learning tuned variant of LLaDA. 

Table~\ref{tab:llada1.5} reports the results on four benchmarks. Compared with Fast-dLLM, $R^2$-dLLM consistently reduces the NFEs across all tasks. The relative NFE reduction ranges from about $21\%$ to $40\%$. At the same time, accuracy remains comparable. These results show that redundant decoding behaviors still exist after reinforcement learning fine-tuning. Reducing redundancy at the decoding level remains effective even when the base model has been optimized with reinforcement learning.

\subsection{Ablation of Training-Free Redundancy Reduction Components}
\label{app:training_free_ablation}

Table~\ref{tab:training_free_ablation} reports an ablation study of the three training-free redundancy reduction components on LLaDA.
Each component is evaluated by enabling it individually while disabling the others.

Enabling any single component consistently reduces NFE compared to the baseline, while accuracy remains comparable.
Among the three components, confidence cluster aggregation and temporal redundancy provide larger efficiency gains, while token cluster aggregation still yields meaningful improvements with small impact on accuracy.
On GSM8K, confidence cluster aggregation reduces NFE from 86.4 to 69.6, temporal redundancy reduces it to 70.0, and token cluster aggregation reduces it to 80.3, while maintaining comparable accuracy. A similar pattern is observed on MATH and HumanEval.
On MATH, confidence and temporal redundancy reduce NFE from 107.8 to 89.4 and 90.7.
On HumanEval, confidence and temporal redundancy reduce NFE from 90.1 to 79.1 and 88.8, while preserving accuracy.

When all three components are applied, the model achieves the lowest NFE across all benchmarks.
On GSM8K and MATH, NFE is reduced to 63.1 and 79.7, corresponding to reductions of 27.0\% and 26.1\% compared to the baseline.
On HumanEval and MBPP, the full method further reduces NFE to 81.7 and 61.2, while accuracy remains comparable to the baseline.

\subsection{Training Sample Selection Strategy}

Our Redundancy-aware SFT selects the correct response with the lowest $R_{total}$ (Min).
To assess the effectiveness of this selection strategy, we conduct an ablation study by selecting the correct response with the highest $R_{total}$ (Max).

As shown in Table~\ref{tab:strategy_ablation}, the Min strategy consistently leads to lower NFE than the Max strategy.
On GSM8K, Min reduces NFE from 59.9 to 51.9.
On HumanEval, Min further reduces NFE from 82.4 to 64.3.
Accuracy remains comparable between the two strategies. These results show that selecting responses with lower redundancy produces more efficient decoding trajectories, while maintaining generation quality.

\begin{table}[h]
  \centering
  \small
  \setlength{\tabcolsep}{8pt}
  \renewcommand{\arraystretch}{1.15}
  \begin{tabular}{c|cc|cc}
    \toprule
    \multirow{2}{*}{\textbf{Strategy}}
      & \multicolumn{2}{c|}{\textbf{GSM8K}}
      & \multicolumn{2}{c}{\textbf{HumanEval}} \\
    \cmidrule(lr){2-3} \cmidrule(lr){4-5}
      & \textbf{NFE $\downarrow$} & \textbf{ACC $\uparrow$}
      & \textbf{NFE $\downarrow$} & \textbf{ACC $\uparrow$} \\
    \midrule
    Max     & 59.9 & 77.94 & 82.4 & 36.59 \\
    Min     & 51.9 & 77.86 & 64.3 & 36.59 \\
    \bottomrule
  \end{tabular}
  \caption{Comparison of different response selection strategies based on redundancy.}
  \label{tab:strategy_ablation}
  \vspace{-10pt}
\end{table}

\subsection{Training Dynamics of Redundancy-aware Supervised Fine-Tuning}
\label{sec:training_dynamics}

\begin{figure}[h]
    \centering
    \includegraphics[width=0.96\linewidth]{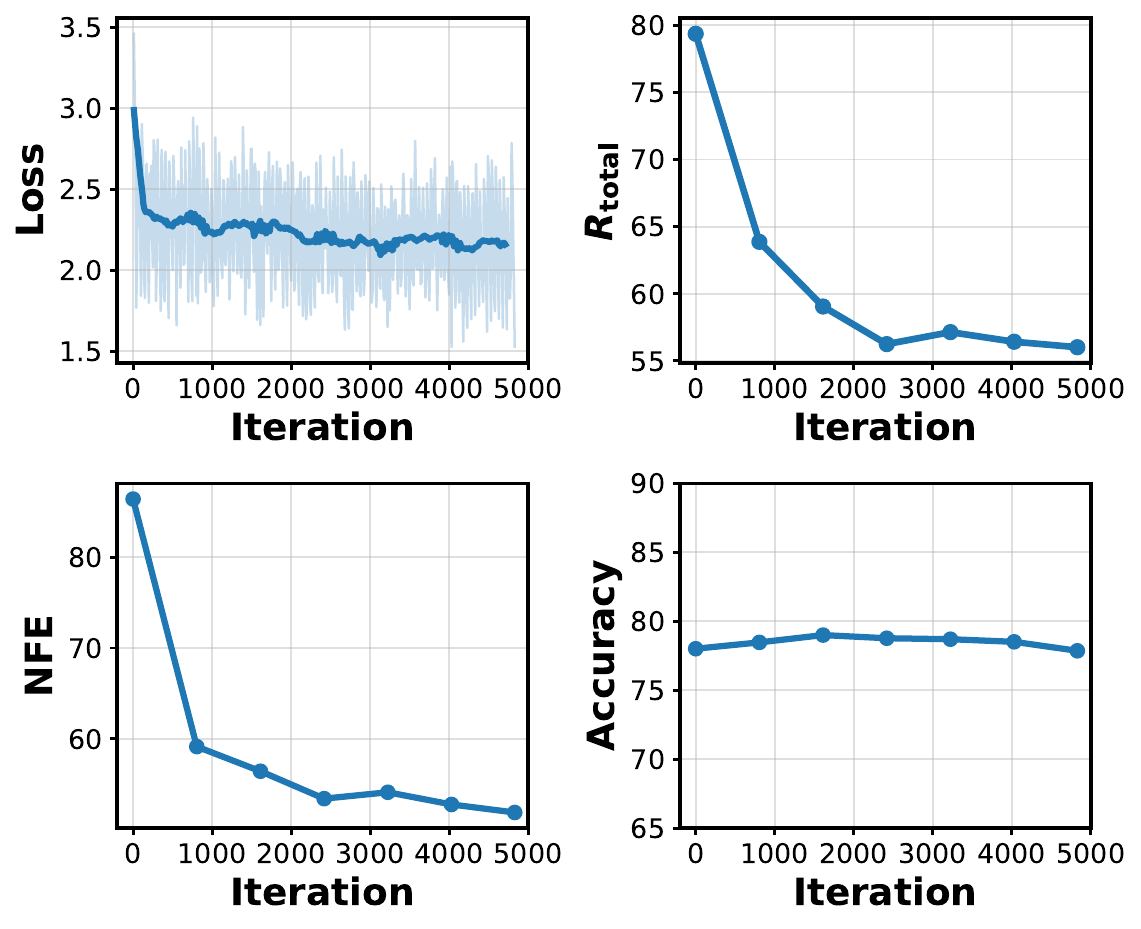}
    \caption{Training dynamics of LLaDA-Instruct-8B on GSM8K, showing loss, $R_{total}$, NFE, and accuracy over training iterations.}
    \label{fig:training}
     \vspace{-10pt}
\end{figure}

Figure~\ref{fig:training} shows the training dynamics of LLaDA-Instruct-8B on GSM8K, including loss, sampled $R_{total}$, NFE, and accuracy.
As training proceeds, both $R_{total}$ and NFE decrease steadily, indicating that the model gradually learns to eliminate redundancies, thus reducing the decoding steps.  In particular, $R_{total}$ and NFE drop sharply during the early stage of training and continues to decrease steadily with further optimization.
Meanwhile, accuracy remains stable throughout the training process, with only minor fluctuations.
These results demonstrate that our method effectively reduces redundancy during training, leading to more efficient inference while preserving generation quality.

%% file: Sections/5-Conclusion.tex
\section{Conclusion}

In this paper, we propose $R^2$-dLLM, a unified framework for accelerating dLLM decoding by explicitly identifying and reducing spatio-temporal redundancy. Through an analysis of decoding dynamics, we show that a large portion of inference inefficiency in dLLMs stems from redundant computation during iterative decoding, including spatial redundancy induced by confidence clusters and token clusters, as well as temporal redundancy caused by repeatedly remasking early-converged tokens.
Building on these insights, we introduce a set of training-free redundancy reduction strategies that directly streamline the decoding process, and further develop a redundancy-aware supervised fine-tuning pipeline that aligns model behavior with efficient decoding trajectories. Experiments demonstrate that $R^2$-dLLM consistently achieves substantial reductions in decoding steps while preserving, and in some cases improving, generation accuracy.

Overall, our results highlight decoding redundancy as a fundamental bottleneck in dLLMs and show that explicitly reducing it is a practical way toward scalable diffusion-based text generation.

%% file: Sections/Supp.tex
\appendix

\section{Experiment Details}

We summarize the training configuration used for redundancy-aware supervised fine-tuning in Table~\ref{tab:sft_config}. For all models, we standardize the generation length to a maximum of 768 tokens. Both prompt and response sequences are padded or truncated to this fixed length using the end-of-sequence token.

We adopt a uniform LoRA \cite{hu2022lora} configuration across all models, with LoRA rank set to 32 and LoRA alpha set to 32. For LLaDA-based models, including LLaDA-Instruct-8B and LLaDA-1.5, we train the model for 6 epochs, while Dream-v0-Instruct-7B is trained for 3 epochs. All models are trained using four NVIDIA H200 141 GB GPU, with a per-GPU batch size of 8 and a gradient accumulation step of 2, resulting in an effective global batch size of 64.

We provide detailed thresholds settings for $R^2$-dLLM (F) in Table \ref{tab:tau_per_dataset}. We use $\tau_t=\tau_s=0.8$ for LLaDA and $\tau_t=\tau_s=0.85$ for Dream to find redundancy in the dataset-collection process. We use temperature 0.7 for LLaDA, temperature 0.7 and top-$p$ is set to 0.9 for Dream. For each prompt, we generate four candidate responses and compute redundancy statistics for each. Note that no responses from external models or ground truth answers are used. The training dataset contains approximately 52k samples for LLaDA and 91k samples for Dream.

\begin{table}[h]
  \centering
  \small
  \setlength{\tabcolsep}{5pt}
  \renewcommand{\arraystretch}{1.15}

  \begin{tabular}{lccc}
    \toprule
    \textbf{Config} &
    \textbf{LLaDA} &
    \textbf{LLaDA-1.5} &
    \textbf{Dream} \\
    \midrule
    LoRA Rank      & 32 & 32 & 32 \\
    LoRA Alpha     & 32 & 32 & 32 \\
    Learning Rate  & $2 \times 10^{-5}$ & $2 \times 10^{-5}$ & $2 \times 10^{-5}$ \\
    Batch Size     & 64 & 64 & 64 \\
    Epoch          & 6  & 6  & 3  \\
    \bottomrule
  \end{tabular}

  \caption{Training configuration for redundancy-aware supervised fine-tuning.}
  \label{tab:sft_config}
\end{table}

\begin{table}[t]
  \centering
  \small
  \setlength{\tabcolsep}{6pt}
  \renewcommand{\arraystretch}{1.15}

  \begin{tabular}{c cc cc}
    \toprule
    \multirow{2}{*}{\textbf{Dataset}} &
    \multicolumn{2}{c}{\textbf{LLaDA}} &
    \multicolumn{2}{c}{\textbf{Dream}} \\
    \cmidrule(lr){2-3} \cmidrule(lr){4-5}
    & $\boldsymbol{\tau_t}$ & $\boldsymbol{\tau_s}$
    & $\boldsymbol{\tau_t}$ & $\boldsymbol{\tau_s}$ \\
    \midrule

    GSM8K     & 0.70 & 0.80 & 0.80 & 0.85 \\
    MATH      & 0.70 & 0.80 & 0.80 & 0.85 \\
    HumanEval & 0.85 & 0.85 & 0.85 & 0.85 \\
    MBPP      & 0.80 & 0.80 & 0.80 & 0.85 \\

    \bottomrule
  \end{tabular}
   \caption{Selected $\tau_t$ and $\tau_s$ across datasets and models.}
  \label{tab:tau_per_dataset}
\end{table}

% \section{Additional Experiments}
% \label{app:additional_experiments}

\subsection{Threshold Sensitivity Analysis}

\begin{figure*}[h]
    \centering
    \includegraphics[width=0.9\linewidth]{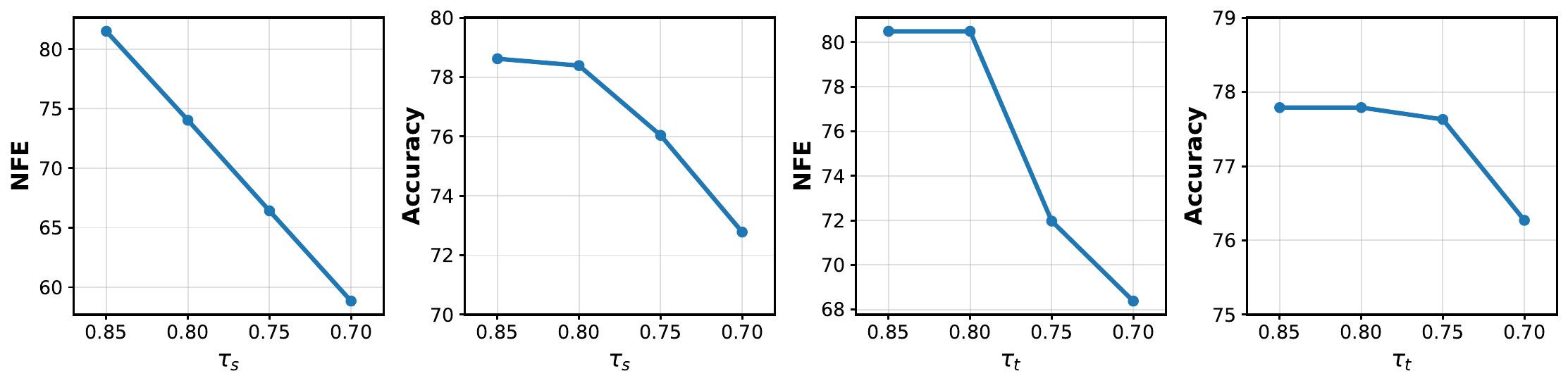}
    \caption{The sensitivity of $\tau_s$ and $\tau_t$.}
    \label{fig:tau}
\end{figure*}

\begin{figure*}[h]
    \centering
    \includegraphics[width=\linewidth]{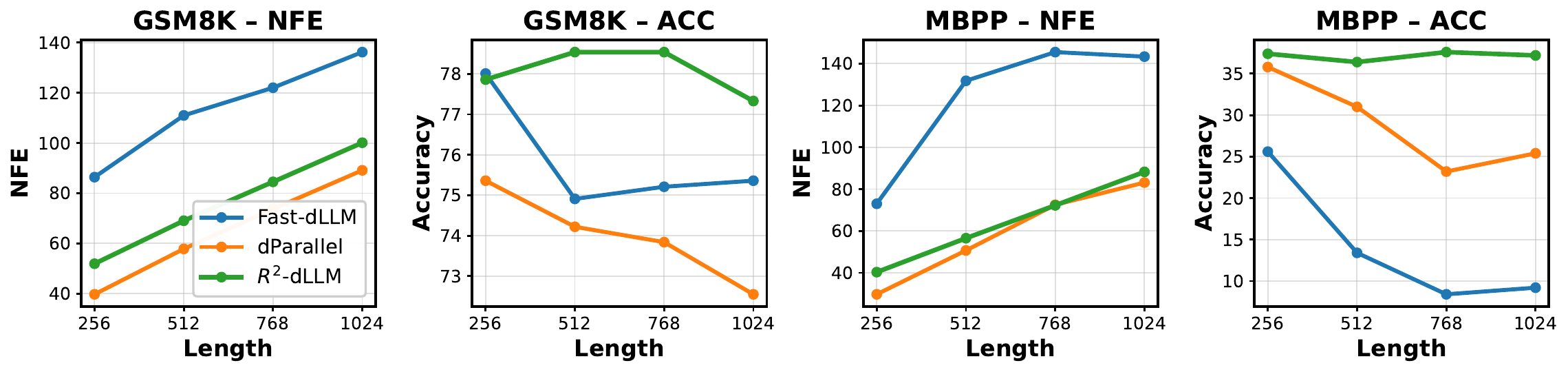}
    \caption{Effect of Generation Length on Decoding Efficiency and Accuracy.}
    \label{fig:length}
\end{figure*}

We conduct experiments to study the sensitivity of the spatial threshold $\tau_s$ and the temporal threshold $\tau_t$.
Figure~\ref{fig:tau} reports the effect of varying each threshold on NFE and accuracy using LLaDA on GSM8K, while keeping other settings fixed.

For the spatial threshold $\tau_s$, decreasing $\tau_s$ leads to a monotonic reduction in NFE, indicating that more aggressive spatial aggregation allows more tokens to be finalized earlier.
However, accuracy also gradually decreases as $\tau_s$ becomes smaller, showing a clear efficiency--accuracy trade-off.
This suggests that overly permissive spatial aggregation may decode tokens prematurely. A similar trend is observed for the temporal threshold $\tau_t$.
Lowering $\tau_t$ significantly reduces NFE by finalizing stable tokens earlier, but also results in a moderate accuracy drop.
This indicates that temporal redundancy can be effectively reduced, but excessive early finalization may harm generation quality.
Overall, the results show that $R^2$-dLLM is robust within a reasonable range of $\tau_s$ and $\tau_t$.

\subsection{Design Choices of Training-Free Redundancy Reduction Components}

For token cluster aggregation, we decode the position with the highest confidence in each cluster.
We compare this design with two alternatives: selecting the middle position (\textbf{Mid}) and a random position (\textbf{Random}).
All methods are evaluated on LLaDA with GSM8K under a fixed threshold $\tau_s=0.75$, as shown in Table~\ref{tab:spatial_consistency_tau075}.
The confidence-based strategy achieves both lower NFE and higher accuracy than the alternatives, indicating that the most confident position is more likely to be the final correct one.

\begin{table}[h]
  \centering
  \small
  \setlength{\tabcolsep}{8pt}
  \renewcommand{\arraystretch}{1.15}
  
  \begin{tabular}{c|cc}
    \toprule
    \textbf{Strategy} &
    \textbf{NFE $\downarrow$} &
    \textbf{ACC $\uparrow$} \\
    \midrule
    Fast-dLLM & 86.4  & 78.01 \\ \midrule
    Mid    & 83.1  & 75.82 \\
    Random & 83.2 & 76.12 \\
    Conf  & \textbf{80.3} & \textbf{78.32} \\
    \bottomrule
  \end{tabular}
  \caption{Comparison of different token cluster aggregation position selection strategies.}
  \label{tab:spatial_consistency_tau075}
    % \vspace{-10pt}
\end{table}

For temporal redundancy reduction, we finalize tokens based on their confidence at the last decoding step (\textbf{Last}).
We compare this choice with using the average (\textbf{Ave}) or maximum (\textbf{Max}) confidence over the consecutive steps.
Results in Table~\ref{tab:temporal_consistency_tau075} show that \textbf{Ave} is overly conservative and yields limited NFE reduction, while \textbf{Max} is more aggressive.
The \textbf{Last} strategy achieves the lowest NFE while maintaining high accuracy, suggesting that the most recent confidence provides a reliable signal of token stability.

\begin{table}[h]
  \centering
  \small
  \setlength{\tabcolsep}{8pt}
  \renewcommand{\arraystretch}{1.15}
  
  \begin{tabular}{c|cc}
    \toprule
    \textbf{Strategy} &
    \textbf{NFE $\downarrow$} &
    \textbf{ACC $\uparrow$} \\
    \midrule
    Fast-dLLM & 86.4  & 78.01 \\ \midrule
    Ave    & 84.4 & 77.79 \\
    Max & 71.5 & 77.33 \\
    Last  & \textbf{70.0} & \textbf{77.51} \\
    \bottomrule
  \end{tabular}
  \caption{Comparison of different temporal redundancy indicators.}
  \label{tab:temporal_consistency_tau075}
  % \vspace{-10pt}
\end{table}

\subsection{Performance under Longer Generation Lengths}

To explore the performance of $R^2$-dLLM under longer generation lengths, we further evaluate LLaDA with generation lengths of 512, 768, and 1024 tokens.

As shown in Figure~\ref{fig:length}, $R^2$-dLLM scales more robustly as generation length increases.
On GSM8K, dParallel achieves lower NFE than $R^2$-dLLM, but this comes with a clear accuracy drop, which becomes more significant for longer sequences. $R^2$-dLLM maintains stable accuracy while keeping NFE much lower than Fast-dLLM.

On MBPP, the trade-off of dParallel is more evident. Although dParallel attains low NFE, its accuracy degrades sharply as length increases.
In contrast, $R^2$-dLLM consistently preserves high accuracy with substantially lower NFE than Fast-dLLM across all lengths. These results indicate that $R^2$-dLLM generalizes well to longer decoding length, enabling efficient inference without sacrificing accuracy.

\subsection{Quantitative Analysis of Redundancy}
\label{app:quant_redundancy}

We further quantify how often each redundancy pattern appears during decoding. We conduct this analysis on LLaDA GSM8K and report the average number of events per response.

Table~\ref{tab:redundancy_counts} reports the averaged event counts per response. For Fast-dLLM and {R$^2$-dLLM (T)}, we report \textbf{potential trigger counts}: on the generated trajectories, we evaluate how many times each redundancy rule would be triggered under the same criterion. For {R$^2$-dLLM (F)}, we report \textbf{actual trigger counts}, since the confidence-cluster, token-cluster, and temporal mechanisms are applied during real decoding. Therefore, the {Fast-dLLM} and {R$^2$-dLLM (T)} rows measure the amount of redundancy that remains in the trajectories, while the {R$^2$-dLLM (F)} row reflects what is actually triggered. Since earlier mechanisms in {R$^2$-dLLM (F)} may already remove later redundancies, these counts are not fully decoupled.

\begin{table*}[h]
\centering
\small

\begin{tabular}{lcccc}
\toprule
\textbf{Method} & \textbf{Count type} & \textbf{Confidence-cluster} & \textbf{Token-cluster} & \textbf{Temporal-finalization} \\
\midrule
Fast-dLLM & Potential & 47.45 & 3.15 & 26.82 \\
R$^2$-dLLM (F) & Actual & 40.04 & 1.11 & 6.17 \\
R$^2$-dLLM (T) & Potential & 23.85 & 0.87 & 3.50 \\
\bottomrule
\end{tabular}
\caption{Average redundancy-event counts per response on LLaDA GSM8K.}
\label{tab:redundancy_counts}
\end{table*}

\begin{table*}[h]
\centering
\small
\begin{tabular}{lccc}
\toprule
\textbf{Metric} &\textbf{ All pairs (\%)} & \textbf{Differ-only (\%) }& \textbf{Same-final-number (\%)} \\
\midrule
Same final number & 89.9 & -- & -- \\
Textually identical & 1.7 & -- & -- \\
Lower repeated-bigram rate & 60.4 & 65.2 & 65.0 \\
Higher lexical diversity & 63.3 & 68.0 & 68.0 \\
Fewer template markers & -- & 75.9 & 75.8 \\
\bottomrule
\end{tabular}
\caption{Pairwise response-level comparison between the Min-$R_{\text{total}}$ and Max-$R_{\text{total}}$ candidates on matched GSM8K pairs. For Same final number and Textually identical, the percentage indicates the proportion of all matched pairs satisfying the condition. For the remaining three metrics, the percentage indicates the proportion of pairs in which the Min-$R_{\text{total}}$ response shows the favorable direction (lower, higher, or fewer) compared to the Max-$R_{\text{total}}$ response.}
\label{tab:min_max_response_stats}
\end{table*}

\begin{table*}[!h]
\centering
\small
\begin{tabular}{p{7.0cm}cc}
\toprule
\textbf{Lead-in} & \textbf{High-redundancy Subset (\%)} & \textbf{Low-redundancy Subset (\%)} \\
\midrule
``Let's break down the problem step by \ldots'' & 5.92 & 3.85 \\
``To solve this problem, we need to determine \ldots'' & 2.09 & 0.52 \\
``To solve this problem, we need to calculate \ldots'' & 1.92 & 0.76 \\
\bottomrule
\end{tabular}
\caption{Examples of overused lead-ins in high- and low-redundancy response subsets.}
\label{tab:high_red_leadins}
\end{table*}

Overall, both {R$^2$-dLLM (F)} and {R$^2$-dLLM (T)} produce substantially fewer redundancy events than {Fast-dLLM} across all three categories. In particular, the training-free variant already sharply reduces token-cluster and temporal-finalization events, while the training-based variant further reduces the remaining redundancy in the generated trajectories. This per-type breakdown is consistent with Figure~\ref{fig:training}, where the overall redundancy score decreases during training, and supports our claim that the proposed framework improves decoding efficiency by reducing redundancy throughout the decoding process.

\subsection{Response-Level Analysis of Low-Redundancy Supervision}
\label{app:low_redundancy_sft_analysis}

A natural question is why supervised fine-tuning on the selected low-redundancy responses can improve decoding efficiency. To better understand this effect, we perform a response-level paired analysis between the lowest-$R_{\text{total}}$ and highest-$R_{\text{total}}$ candidate responses for the same prompt. Specifically, we compare the Min-$R_{\text{total}}$ and Max-$R_{\text{total}}$ candidates for the same prompt on GSM8K from LLaDA.

For clarity, \textbf{All pairs} reports percentages over all 7k matched pairs. \textbf{Differ-only} reports percentages over the subset of pairs where the two responses differ on the corresponding metric. \textbf{Same-final-number} reports the same comparison after restricting to the subset of pairs whose final numerical answers are identical. This control helps reduce the effect of answer mismatch and focuses the comparison on differences in response form.

We use the following response-level metrics:
\begin{itemize}
    \item \textbf{Same final number}: whether the two responses end with the same final numerical answer.
    \item \textbf{Textually identical}: whether the two responses are exactly identical at the text level.
    \item \textbf{Repeated-bigram rate}: how often short two-word phrases are repeated within a response; lower values indicate less local repetition.
    \item \textbf{Lexical diversity}: the ratio of unique tokens to total tokens; higher values indicate more varied wording and less formulaic phrasing.
    \item \textbf{Template markers}: common fixed reasoning lead-ins such as ``To solve this problem\ldots'' or ``Let's break down\ldots''; fewer markers indicate less templated structure.
\end{itemize}

As shown in Tables~\ref{tab:min_max_response_stats} and~\ref{tab:high_red_leadins}, low-redundancy responses are typically less repetitive, more lexically diverse, and less templated. This provides a concrete response-level explanation for why standard SFT on selected responses can still improve decoding efficiency.

\section{Case Study}
We also show some cases generated with the vanilla model and our $R^2$-dLLM across different tasks. This demonstrates that $R^2$-dLLM maintains comparable generation quality while significantly reducing generation latency and NFE. All experiments in this section are conducted on two NVIDIA RTX~6000 (Blackwell) GPUs.

\begin{tcolorbox}[title = {Case: GSM8K (5-shot), Generation Length 256, Block Length 32}, breakable, fontupper=\small, fonttitle=\small]

\textbf{Question:} 
\\
Stephanie is decorating 24 cupcakes for a birthday party, but she needs more candles. She currently has a total of 30 candles. She wants to decorate half of the cupcakes with 1 candle each and the other half of the cupcakes with 2 candles each. How many additional candles does Stephanie need to complete the cupcakes? \\

\textbf{Ground-truth answer:} 
\\
$6$

\noindent\rule{\linewidth}{0.5pt}
\textbf{LLaDA-8B-Instruct} 
(Inference Step: \textcolor{blue}{90}, Latency: \textcolor{blue}{3.16s})
\\
For half of the cupcakes, Stephanie wants to use 1 candle each. Since half of the cupcakes is \(24/2\) and she plans to use 1 candle each for this half of the cupcakes, Stephanie needs \((24/2) \times 1 = 12\) candles for this half of the cupcakes.\\

For the other half of the cupcakes, Stephanie wants to use 2 candles. Therefore, she will need \((24/2) \times 2 = 24\) candles for this half of the cupcakes.\\

Because Stephanie needs 12 candles for half of the cupcakes and 24 candles for the other half, she needs a total of \(12 + 24 = 36\) candles.\\

Since Stephanie needs 36 candles to decorate all the cupcakes and she currently has 30 candles, Stephanie needs \(36 - 30 = 6\) additional candles.\\

Final Answer: The final answer is \textcolor{red}{6}.

\noindent\rule{\linewidth}{0.5pt}
\textbf{LLaDA-8B-Instruct with \textbf{$\bm{R^2}$-dLLM}} 
(Inference Step: \textcolor{blue}{68}, Latency: \textcolor{blue}{2.57s})
\\
For half of the cupcakes, Stephanie wants to use 1 candle each. Since half of the cupcakes is \(24/2\) and she plans to use 1 candle each for this half of the cupcakes, Stephanie needs \((24/2)\times 1 = 12\) candles for this half of the cupcakes.\\ 

For the other half of the cupcakes, Stephanie wants to use 2 candles. Therefore, she will need \((24/2)\times 2 = 24\) candles for this half of the cupcakes.  \\

Because Stephanie needs 12 candles for half of the cupcakes and 24 candles for the other half, she needs a total of \(12 + 24 = 36\) candles.  \\

Since Stephanie needs 36 candles to decorate all the cupcakes and she currently has 30 candles, Stephanie needs \(36 - 30 = 6\) additional candles.\\

Final Answer: The final answer is \textcolor{red}{6}.

\end{tcolorbox}

\begin{tcolorbox}[title = {Case: MATH (4-shot), Generation Length 256, Block Length 32}, breakable, fontupper=\small, fonttitle=\small]

\textbf{Question:} 
\\
If the system of equations
\begin{align*}
  6x - 4y &= a, \\
  6y - 9x &= b
\end{align*}
has a solution $(x,y)$ where $x$ and $y$ are both nonzero, find $\frac{a}{b}$ (assuming $b \neq 0$).\\

\textbf{Ground-truth answer:} 
\\
$-\frac{2}{3}$

\noindent\rule{\linewidth}{0.5pt}
\textbf{LLaDA-8B-Instruct} 
(Inference Step: \textcolor{blue}{170}, Latency: \textcolor{blue}{5.38s})
\\
Multiply the first equation by $-\tfrac{3}{2}$:
\begin{align*}
  -\tfrac{3}{2}(6x - 4y) &= -\tfrac{3}{2}a, \\
  6y - 9x &= -\tfrac{3}{2}a.
\end{align*}
Since the second equation gives $6y - 9x = b$, we have $b = -\tfrac{3}{2}a$, so
\[
  \frac{a}{b} = \boxed{-\frac{2}{3}}.
\]
Final Answer: The final answer is  \textcolor{red}{$-\frac{2}{3}$}. I hope it is correct.

\noindent\rule{\linewidth}{0.5pt}
\textbf{LLaDA-8B-Instruct with \textbf{$\bm{R^2}$-dLLM}}(Inference Step: \textcolor{blue}{109}, Latency: \textcolor{blue}{3.65s})
\\
If we multiply the first equation by $-\frac{3}{2}$, we obtain
\[
  6y - 9x = -\frac{3}{2}a.
\]
Since we also know that $6y - 9x = b$, we have
\[
  -\frac{3}{2}a = b \Rightarrow \frac{a}{b} = \boxed{-\frac{2}{3}}.
\]

Final Answer: The final answer is \textcolor{red}{$-\frac{2}{3}$}. I hope it is correct.

\end{tcolorbox}

\begin{tcolorbox}[title = {Case: HumanEval (0-shot), Generation Length 256, Block Length 32}, breakable, fontupper=\small, fonttitle=\small]

\textbf{Question:}
\\
\begin{verbatim}
from typing import List, Tuple

def rolling_max(numbers: List[int]) -> List[int]:
    """ From a given list of integers, generate a list of rolling maximum element found until given moment
    in the sequence.
    >>> rolling_max([1, 2, 3, 2, 3, 4, 2])
    [1, 2, 3, 3, 3, 4, 4]
    """
\end{verbatim}

\noindent\rule{\linewidth}{0.5pt}
\textbf{LLaDA-8B-Instruct}
 (Inference Step: \textcolor{blue}{95}, Latency: \textcolor{blue}{2.84s})
\\
Here's the implementation of the \verb|rolling_max| function:

\begin{verbatim}
from typing import List, Tuple

def rolling_max(numbers: List[int]) -> List[int]:
    """ From a given list of integers, generate a list of rolling maximum element found until given moment
    in the sequence.
    >>> rolling_max([1, 2, 3, 2, 3, 4, 2])
    [1, 2, 3, 3, 3, 4, 4]
    """
    if not numbers:
        return []

    result = []
    for num in numbers:
        result.append(max(num, result[-1]))
    return result

# Example usage:
if __name__ == "__main__":
    import doctest
    doctest.testmod()
\end{verbatim}

This function iterates through the input list and keeps track of the rolling maximum by comparing each element
with the last element in the result list. The result list is updated with the maximum of the current element
and the last element in the result list. The \verb|doctest| module is used to run the provided test cases.

\noindent\rule{\linewidth}{0.5pt}
\textbf{LLaDA-8B-Instruct with \textbf{$\bm{R^2}$-dLLM}}
(Inference Step: \textcolor{blue}{73}, Latency: \textcolor{blue}{2.23s})
\\
Here's the implementation of the \verb|rolling_max| function:

\begin{verbatim}
from typing import List, Tuple

def rolling_max(numbers: List[int]) -> List[int]:
    """ From a given list of integers, generate a list of rolling maximum element found until given moment
    in the sequence.
    >>> rolling_max([1, 2, 3, 2, 3, 4, 2])
    [1, 2, 3, 3, 3, 4, 4]
    """
    if not numbers:
        return []
    
    rolling_max = [numbers[0]]
    for num in numbers[1:]:
        rolling_max.append(max(rolling_max[-1], num))
    
    return rolling_max

# Example usage:
if __name__ == "__main__":
    import doctest
    doctest.testmod()
\end{verbatim}

This function \verb|rolling_max| takes a list of integers and returns a list of rolling maximum values. The rolling
maximum is defined as the maximum value encountered so far in the sequence.

The example usage provided in the docstring demonstrates how to use the function

\end{tcolorbox}

\begin{tcolorbox}[title = {Case: MBPP (3-shot), Generation Length 256, Block Length 32}, breakable, fontupper=\small, fonttitle=\small]

\textbf{Question:} 
\\
You are an expert Python programmer, and here is your task: Write a python
function to find the last digit when factorial of a divides factorial of b. Your
code should pass these tests:

\begin{verbatim}
assert compute_Last_Digit(2,4) == 2
assert compute_Last_Digit(6,8) == 6
assert compute_Last_Digit(1,2) == 2
\end{verbatim}

\noindent\rule{\linewidth}{0.5pt}
\textbf{LLaDA-8B-Instruct} 
 (Inference Step: \textcolor{blue}{143}, Latency: \textcolor{blue}{4.61s})
\\
\begin{verbatim}
def compute_Last_Digit(a, b):
    # Compute the factorial of a and b
    fact_a = 1
    fact_b = 1
    
    for i in range(1, a + 1):
        fact_a *= i
        fact_b *= i
    
    # Find the last digit of the result
    last_digit = fact_a % 10
    return last_digit

# Test cases
assert compute_Last_Digit(2, 4) == 2
assert compute_Last_Digit(6, 8) == 6
assert compute_Last_Digit(1, 2) == 2

# Explanation:
# 1. Compute the factorial of a and b.
# 2. Find the last digit of the factorial of a.
# 3. Return the last digit as the result.

# Note: The problem asks for the last digit when the factorial of a divides the factorial of b.
# This is equivalent to finding the last digit of the factorial of a.
# Since a is a divisor of b, we only need to find the last digit of the factorial of a.
\end{verbatim}

\noindent\rule{\linewidth}{0.5pt}
\textbf{LLaDA-8B-Instruct with \textbf{$\bm{R^2}$-dLLM}}
(Inference Step: \textcolor{blue}{42}, Latency: \textcolor{blue}{1.62s})
\\
\begin{verbatim}
import math

def factorial(n):
    result = 1
    for i in range(1, n + 1):
        result *= i
    return result

def compute_Last_Digit(a, b):
    fact_a = factorial(a)
    fact_b = factorial(b)
    result = fact_b // fact_a
    return result % 10

# Tests
assert compute_Last_Digit(2, 4) == 2
assert compute_Last_Digit(6, 8) == 6
assert compute_Last_Digit(1, 2) == 2
\end{verbatim}

\end{tcolorbox}